# Automated Translation of Rebar Information from GPR Data into As-Built BIM: A Deep Learning-based Approach

**Zhongming Xiang, Ph.D. Candidate[1], Ge Ou, Assistant Professor[2], and Abbas Rashidi, Assistant Professor [3]**

[1]Dept. of Civil and Environmental Engineering, Univ. of Utah, Salt Lake City, UT 84112; e-mail: zhongming.xiang@utah.edu
[2]Dept. of Civil and Environmental Engineering, Univ. of Utah, Salt Lake City, UT 84112; e-mail: ge.ou@utah.edu
[3]Dept. of Civil and Environmental Engineering, Univ. of Utah, Salt Lake City, UT 84112; e-mail: abbas.rashidi@utah.edu

## ABSTRACT

Building Information Modeling (BIM) is increasingly used in the construction industry, but existing studies often ignore embedded rebars. Ground Penetrating Radar (GPR) provides a potential solution to develop as-built BIM with surface elements and rebars. However, automatically translating rebars from GPR into BIM is challenging since GPR cannot provide any information about the scanned element. Thus, we propose an approach to link GPR data and BIM according to Faster R-CNN. A label is attached to each element scanned by GPR for capturing the labeled images, which are used with other images to build a 3D model. Meanwhile, Faster R-CNN is introduced to identify the labels, and the projection relationship between images and the model is used to localize the scanned elements in the 3D model. Two concrete buildings is selected to evaluate the proposed approach, and the results reveal that our method could accurately translate the rebars from GPR data into corresponding elements in BIM with correct distributions.

## INTRODUCTION

Due to the significant technological improvement in the visualization, automation, efficiency, and productivity in all lift cycles of building management, building information modeling (BIM) has been widely applied to construction projects (Zhao, 2017). As a result, the investment in BIM-related products is sustainably growing in recent years, with an annual increase rate of 15.5% (ValuatesReports, 2020). To generate as-built BIMs, vendors have launched various products (i.e., Revit®, SketchUp®, MicroStation®, etc.), which utilize the geometrical information of building components as the inputs (Wang et al., 2015). For a building with sufficient drawings, it is efficient to adopt these products to develop as-built BIMs as the geometries are available from the drawings. However, for some buildings lacking drawings, the commercial tools might be inefficient due to the implementation of capturing component sizes, which are time-consuming and error-prone. Thus, finding a reliable solution to generate BIMs for as-built buildings is still an ongoing academic exploration.

BIM is typically indicated as the 3D model with identified categories of building elements, and most of the studies are focused on how to generate as-built BIMs for visible elements, including shear walls, slabs, columns, doors, and so on. However, one of the embedded components, rebar, is usually ignored in the process of BIM development due to the inaccessibility of rebar. This phenomenon has limited the BIM applications, such as structural health monitoring maintenance and building renovation. Thus, it is necessary to develop a complete BIM with surface elements and embedded rebars for as-built buildings. To accomplish this goal, three major challenges should be tackled: 1) developing an accurate BIM model for surface elements, 2) localizing rebars from concrete elements, and 3) translating rebars into corresponding elements into BIM.

To address the first challenge in BIM development for surface elements, two sub-steps are involved. One is 3D data (point cloud) acquisition from an infrastructure. Researchers have proposed various approaches, which can be categorized as image-based methods (Rashidi and Karan, 2018) and time-of-flight (ToF)-based methods (Turkan et al., 2013). Another sub-step is to identify elements from a point cloud. To address this issue, several kinds of methods have been explored based on the projection relationship between 2D images and 3D models (Braun and Borrmann, 2019), placement rules of structural elements (e.g., columns are vertically placed.) (Lu et al., 2019), or geometrical relationships between different elements (e.g., windows are normally placed on the walls and higher than doors) (Wang et al., 2015).

For rebar localization, several non-destructive testing (NDT) tools [i.e., ground penetrating radar (GPR) (Lai et al., 2018), impact echo testing (Hong et al., 2015), and ultrasonic testing (Algernon et al., 2011)], are available. GPR is one of the most popular NDT tools and has been widely used to localize rebar due to the advantages of portable size, ease of operation, and high accuracy (Xiang et al., 2019a). Since rebar signals are recorded as hyperbolic patterns in GPR data, developing an efficient method to explain the hyperbolas is the major task for rebar localization. As a fundamental step, rebar recognition, or indicated as hyperbolic pattern recognition, has been well studied, and researchers have proposed various methods, which can be classified as pattern-based methods (Xiang et al., 2021) and machine learning-based methods (Lei et al., 2019). The subsequent step is rebar depth and size determination, which involves 1) modifying GPR device itself (Agred et al., 2018), 2) combining GPR device with other NDT tools (Wiwatrojanagul et al., 2018), and 3) utilizing advanced image processing techniques (Xiang et al., 2020).

For the rebar insertion in the as-built BIM, however, there is not any effective study available yet. The limited publications are about placing rebars into BIM based on the information from 2D drawings (Choi et al., 2014). On the commercial side, some vendors have designed GPR devices that can generate 3D rebars, but there are two obvious drawbacks: manual calibration during scanning and no guaranteed compatibility with BIM software (Lai et al., 2018). Therefore, inserting identified rebars into the corresponding element in an as-built BIM is almost a blank area that needs a systematic solution. To solve this issue, this research proposes a deep learning approach to automatically translate rebars from GPR data into an as-built BIM based on the photogrammetry method of 3D reconstruction.

Since rebars in GPR data and the concrete elements in BIM are generated from two systems, to translate rebars into BIM, mapping information should be introduced to integrate these two domains. Thus, after scanning each element with the GPR device and capturing images of the elements from different angles and positions, this research designs a label that is attached to each element one by one, and one image (called labeled image) is captured for each element with the attached label. Then, these labeled images are used together with other images to reconstruct a 3D model. After that, this research introduces a deep learning method, Faster RCNN, to implement the automated localization of the labels among all 2D images. The label locations in the 2D images are then mapped to the 3D model based on the projection relationship between 2D images and the 3D model. Subsequently, the scanned elements can be identified in a 3D model, which means the localized rebars in the GPR data are linked to the corresponding element in the BIM model. More details about the proposed approach will be discussed in the following sections. Meanwhile, two concrete buildings are selected as testbeds to demonstrate the feasibility of the proposed approach.

**RESEARCH METHODOLOGY**

The framework of the proposed approach is presented in **Figure 1**, which depicts the components of developing a BIM model based on images, recognizing labels through Faster RCNN, projecting identified labels from 2D images to the 3D model, localizing rebars in GPR data, and linking rebars in GPR data to corresponding elements in the 3D model. In this framework, there are two fundamental steps: BIM development and rebar localization, which, however, will not be discussed in this research since there are already mature solutions for these steps.

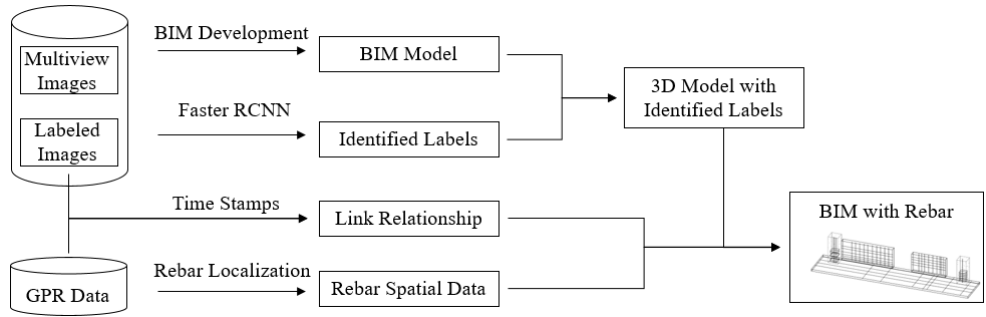

**Figure 1. The framework of the proposed approach**

On the one hand, as a well-established technique, photogrammetry has shown significant efficiency to reconstruct a 3D model with low cost but high quality (Golparvar-Fard et al., 2011). Several mature commercial programs can implement this task. In this research, the ContextCapture, a reliable software produced by Bentley, is used to reconstruct a 3D model to save program settings. Meanwhile, this research does not explore new techniques for element identification from point clouds and directly assumes that the elements (i.e., columns, shear walls, and slabs) are identified.

On the other hand, rebar localization in GPR data includes two sub-steps: rebar recognition and rebar depth and size determination. For rebar recognition, the authors have proposed a method that utilized the distinguish features of rebar signals in the frequency domain to differentiate the rebar from the background based on a series of frequency filters (Xiang et al., 2021). For rebar depth determination, the inverse correlation between the maximum intensity in GPR data and the travel time of electromagnetic waves is intrinsic and has been used to determine rebar depth in several studies (Dinh et al., 2018; Kaur et al., 2015). For rebar size determination, it is the least reliable application among all kinds of applications of GPR (Bungey 2004). Thus, this research combines the frequency filter method and inverse correlation method to recognize rebar and determine the rebar depth, respectively, while the rebar size is predefined as a uniform diameter of 12.7mm. Therefore, two residual steps, Scanned Element Recognition and Rebar Translation from GPR data into BIM, are critical for completed BIM development and will be the focus of this manuscript.

*Deep Learning-based Scanned Element Recognition*

This research designs a label attached to each element which is scanned by the GPR device (**Figure 2**). Then, an image is captured for each labeled element used with other images to reconstruct a 3D model. The label is first detected in images and then projected to the 3D model. In this way, the scanned elements can be identified in BIM to map the GPR data. The procedure is elaborated in the following two sub-steps:

**A. Label Recognition in 2D:** The Faster R-CNN, designed for objection detection, is introduced here to recognize labels in 2D. Compared with traditional CNN, Faster R-CNN has several technical features, including (1) using selective search to extract thousands of region proposals, (2) converting these region proposals as the feature map of the last convolutional layer, and (3) rescaling all region proposals into a uniform size through a region of interest pooling layer. The Faster RCNN is trained by 300 labeled images and subsequently applied to detect the labels among the images used for 3D reconstruction.

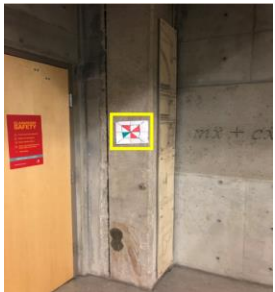
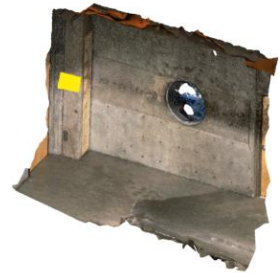

**Figure 2. Labeled image to indicate the element scanned by GPR device**

**Figure 3. Identified label in the 3D model**

**B. Label Projection from 2D to 3D:** The recognized labels in 2D images are then mapped to the 3D model based on the projection relationship between images and the 3D model. The coordinate transformation between the 3D point ($P_3$) and the 2D point ($P_2$) is defined in Equation (1)-(2), where $K$ is the calibration matrix, $R$ is the rotation matrix, and $T$ is the translation matrix. $K$, $R$, and $T$ are camera parameters. As shown in **Figure 3**, the identified label in the wall is mapped into the point cloud at the corresponding location based on the projection relationship.

$$[m_1 \quad m_2 \quad m_3] = K * [R \quad T] * P_3 \quad (1)$$
$$P_2 = [m_1/m_3 \quad m_2/m_3]' \quad (2)$$

*Rebar Translation from GPR data into BIM*

The following step aims to translate rebars from GPR data into the BIM model. **Figure 4a** is an example of GPR data, which contains rebar signals and noise but has no information that can indicate which element the data is scanned from. This phenomenon leads to a problem that there is no discriminative feature to determine which element the rebar is scanned from if there are more than two elements involved in the project. To solve this issue, this research introduces time stamps as the new coordinate to map the GPR data and the scanned elements. Thus, chronological sequences of the GPR data and the labeled images are recorded. The timestamps of the GPR data are $\{(T_1^{Gh}, T_1^{Gv}), (T_2^{Gh}, T_2^{Gv}), ...\}$, where $T_i^{Gh}$ and $T_i^{Gv}$ regard to the two directions of element $E_i$. Meanwhile, the timestamps of the labeled images are $\{T_1^L, T_2^L, ...\}$. Then, we allocate the timestamps as $\{(T_1^{Gh}, T_1^{Gv}) \rightarrow T_1^L, (T_2^{Gh}, T_2^{Gv}) \rightarrow T_2^L, ...\}$ to match each GPR data and labeled image. In this way, the relationship between GPR data and the scanned element is built.

Then, rebars localized from the GPR data should be placed to BIM in corresponding locations. The ratio of rebar locations to the length of GPR data is $r_i = L_i/L_G$, where $L_i$ is the location of the rebar $i$ in the unit of pixel, and $L_G$ is the pixel length of GPR data. Thus, the rebar location in the wall is $r_i * L_w$ (Figure 4b), where $L_w$ is the length of wall. Similarly, rebar depth is determined by $D_r = D_G/H_G * D_{max}$, where $D_G$ is the depth of rebar in the GPR data in the unit of pixel, $H_G$ is the pixel height of the GPR data, and $D_{max}$ is the maximum detection depth of the GPR device. It is worth noting that the GPR data scanned in the vertical direction of the element records rebars in the horizontal direction and vice versa.

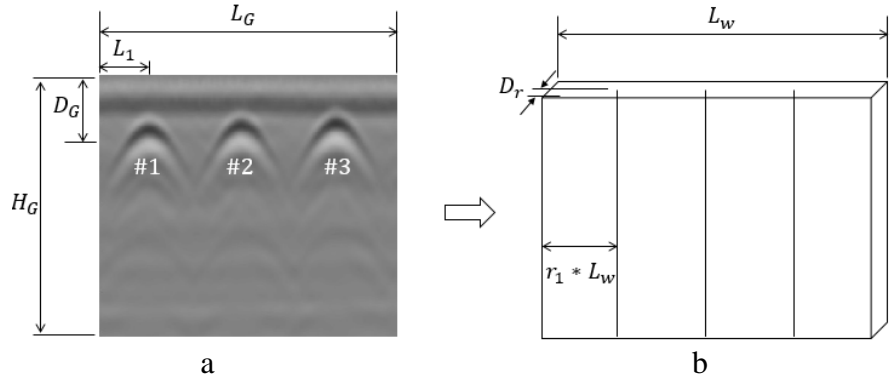

**Figure 4. Rebar translation from GPR data into BIM: (a) GPR data, and (b) BIM model**

**EXPERIMENTAL SETUP AND CASE STUDY**

For validating the performance of the proposed approach, two concrete buildings located at the campus of the University of Utah are selected as the testbeds in this research. The experimental elements of the two buildings are illustrated in **Figure 5**, in which Case 1 involves one shear wall, one column, and one slab, and Case 2 involves two shear walls and one slab. D1 and D2 in Figure 5 denote the two scan directions. To scan rebars, this research adopts the GPR device, Handy Search NJJ-105, which has the maximum detection depth $D_{max}$ of 30 cm and GPR data height $H_G$ of 625 pixels. It should be noted that only the performance of rebar translation will be discussed since the rebar localization and BIM development are not within the scope of this research.

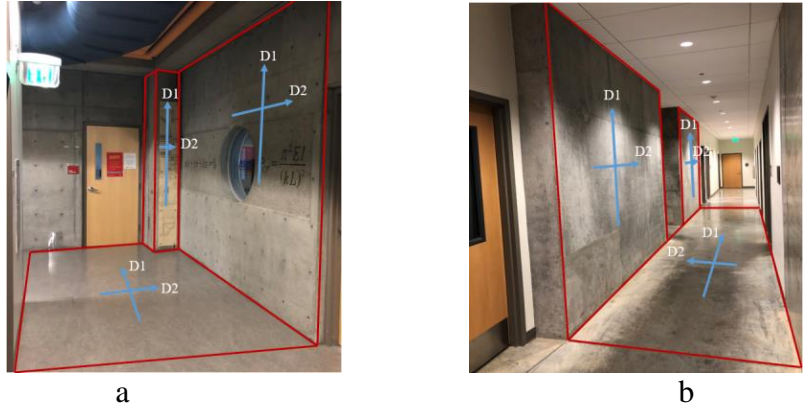
<div style="text-align:center">a            b</div>

**Figure 5. The study regions: (a) Case 1; and (b) Case 2**

Industry Foundation Classes (IFC) is introduced in this research to convert the 3D models into files that are readable in the common BIM platforms (i.e., Revit®). Eventually, 231 images for Case 1 and 296 images for Case 2 are captured to reconstruct 3D models. Among these images, there are three labeled images for Case 1 and three labeled images for Case 2. Similarly, each case has three GPR scans corresponding to each element. After processing these images and GPR data with the proposed approach, the BIM models of the two selected buildings are illustrated in **Figure 6**. Overall, all rebars are translated from GPR data into the BIMs, which indicates the labels are correctly detected and projected into 3D models, and the mapping between the scanned elements and the rebars in the GPR data is built successfully.

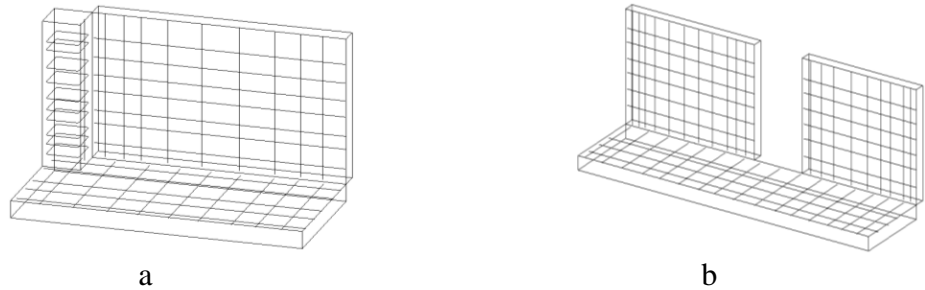
<div style="text-align:center">a            b</div>

**Figure 6. BIMs with surface elements and rebars: (a) Building #1; and (b) Building #2**

The performance of label recognition using Faster RCNN is evaluated first. After training the network, Faster RCNN is utilized to recognize labels among the images that are used to reconstruct 3D models. The accuracy, which denotes if the label is recognized or not, reached 100% for both cases. All labeled images are detected by the Faster RCNN, and the locations of the labels in the images are marked as well. It demonstrates that the deep learning method is applicable for label recognition and feasible to identify the scanned elements. For assessing the performance of the rebar placement, three elements in Case 1 are selected and evaluated. As shown in **Table 1**, all rebars in the GPR data are successfully translated into corresponding elements in the BIM model in both directions, which proves that the proposed method is effective in placing rebars from GPR data into BIMs.

Table 1 Rebar placement in column, wall and slab

| | 3D Model | Section View | GPR Data in Direction 1 | GPR Data in Direction 2 |
|---|---|---|---|---|
| Column | 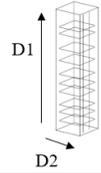 | 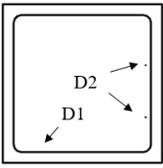 | 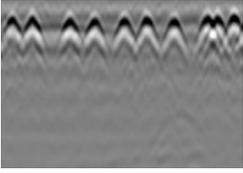 | 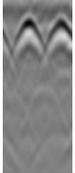 |
| Wall | 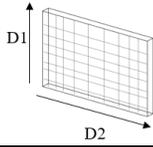 | 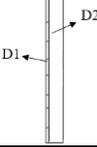 | 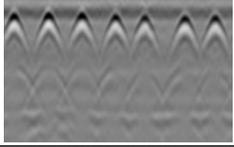 | 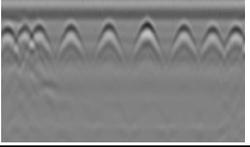 |
| Slab | 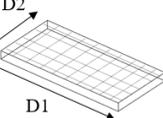 | 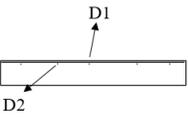 | 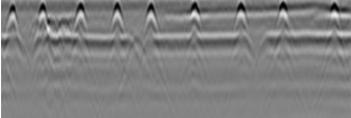 | 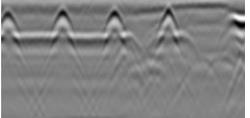 |

## CONCLUSION AND FUTURE WORK

In order to fill the gap of inserting rebars into as-built BIM, this research provided an automated approach that utilized Faster RCNN to translate rebars from GPR data into corresponding elements in BIMs. A predefined image label was attached to the elements that were scanned by the GPR device, and the label was then recognized by the trained Faster RCNN. With timestamps of the GPR data and labeled images, the link between the GPR data and corresponding elements in the BIMs were built, and the rebars were subsequently translated from GPR data into elements. At last, two concrete buildings were selected to validate the feasibility of the proposed approach. According to the results, three conclusions can be summarized:
- The Faster RCNN is a promising method for label recognition as it could detect the existences of all labeled images and localize labels in these images.
- The proposed method could successfully project the labels from 2D images to a 3D model to mark the scanned elements in BIMs.
- The rebars were eventually translated from GPR data into BIMs and correctly placed into corresponding elements.

In this research, a predefined label has been introduced as an extra coordinate to link GPR data and the scanned elements. However, this research has a limitation in that the proposed approach only considered rebars in the second layers. Besides solving this limitation, the authors also plan to explore deep learning methods to build the link based on the pattern features of GPR data and the scanned elements in 2D images. Meanwhile, a comparison discussion of rebar placement will be conducted between the 3D models and the drawings.